\documentclass[conference,onecolumn,onehalfspacing]{IEEEtran}
\IEEEoverridecommandlockouts

\usepackage[utf8]{inputenc}
\usepackage[T1]{fontenc}
\usepackage[english]{babel}
\usepackage{amsmath, amsfonts, amssymb}
\usepackage{algpseudocode}
\usepackage{algorithm}
\usepackage{array}
\usepackage{textcomp}
\usepackage{stfloats}
\usepackage{verbatim}
\usepackage{graphicx}
\usepackage{cite}
\usepackage{import}
\usepackage{bm}
\usepackage{setspace}
\usepackage{dsfont}
\usepackage{indentfirst}
\usepackage[hidelinks]{hyperref}
\usepackage{mathtools}
\usepackage{extarrows}
\usepackage{enumitem}
\usepackage{amsthm}
\usepackage{dsfont}
\usepackage[caption=false]{subfig}
\usepackage[export]{adjustbox}
\usepackage{enumitem}
\usepackage{tcolorbox}
\usepackage[all]{xy}
\usepackage{tabularx,booktabs}
\usepackage{float}
\usepackage{multirow}
\usepackage[switch]{lineno}

\usepackage{tikz}
\usepackage{pgfplots}
\usetikzlibrary{calc}
\pgfplotsset{compat=newest}

\makeatletter 
\@mparswitchfalse%
\makeatother
\reversemarginpar
\usepackage[%
colorinlistoftodos,prependcaption,textsize=tiny
]{todonotes}

\newcommand{\rbrk}[1]{\left( #1 \right)}
\newcommand{\cbrk}[1]{\left\{ #1 \right\}}
\newcommand{\sbrk}[1]{\left[ #1 \right]}

\newcommand{\olsi}[1]{\,\overline{\!{#1}}}

\newcommand{\stopgradient}{\mathrm{sg}}

\newcommand{\mdpstatespace}{S}
\newcommand{\mdpactionspace}{A}
\newcommand{\mdptransition}{T}
\newcommand{\reward}{r}
\newcommand{\mdpdiscount}{\gamma}
\newcommand{\actionpolicy}{\pi}
\newcommand{\state}{s}
\newcommand{\action}{a}
\newcommand{\channel}{g}
\newcommand{\numantennas}{N}
\newcommand{\scheduling}{\alpha}
\newcommand{\power}{\varrho}
\newcommand{\powerbudget}{\varrho_{\text{max}}}
\newcommand{\snr}{\mathrm{SNR}}
\newcommand{\snrtarget}{\overline{\mathrm{SNR}}}
\newcommand{\noisepower}{\sigma}
\newcommand{\longtermreward}{\olsi{R}}
\newcommand{\longtermpower}{\olsi{P}}
\newcommand{\controlnetparameters}{\phi}
\newcommand{\imagerepresentation}{x}
\newcommand{\imageencoder}{e}
\newcommand{\deterministicstate}{h}
\newcommand{\stochasticstate}{z}
\newcommand{\stochasticstatedistribution}{q}
\newcommand{\recurrentfunction}{f}
\newcommand{\distribution}{p}
\newcommand{\predictionhorizon}{H}
\newcommand{\kllossscale}{\beta}
\newcommand{\kllossfactor}{\mu}
\newcommand{\channelencoder}{\zeta}
\newcommand{\channelnetparameters}{\omega}
\newcommand{\channelrepresentation}{c}
\newcommand{\channelpredictor}{\psi}
\newcommand{\channelpredictorhiddenstate}{q}
\newcommand{\actorparameters}{\theta}
\newcommand{\criticparameters}{\xi}
\newcommand{\valuefunction}{v}
\newcommand{\actorentropyscale}{\eta}
\newcommand{\powerpredictorparameters}{\chi}
\newcommand{\truststeps}{\tau}
\newcommand{\consecutivesamples}{\kappa}

\usepackage{soul, xcolor}

\usepackage[acronym]{glossaries}

\begin{document}



\newacronym{ml}{ML}{machine learning}
\newacronym{ai}{AI}{artificial intelligence}
\newacronym{csi}{CSI}{channel state information}
\newacronym{jepa}{JEPA}{joint-embedding predictive architecture}
\newacronym{jea}{JEA}{joint-embedding architecture}
\newacronym{ofdm}{OFDM}{orthogonal frequency division multiplex}
\newacronym{pdr}{PDR}{pedestrian dead reckoning}
\newacronym{rnn}{RNN}{recurrent neural network}
\newacronym{ema}{EMA}{exponential moving average}
\newacronym{mlp}{MLP}{multi-layer preceptron}
\newacronym{gru}{GRU}{gated recurrent unit}
\newacronym{lstm}{LSTM}{long short-term memory}
\newacronym{1nn}{$1$--NN}{nearest neighbor}
\newacronym{adp}{ADP}{angle-delay profile}
\newacronym{wjepa}{W-JEPA}{}
\newacronym{mdp}{MDP}{Markov decision process}
\newacronym{snr}{SNR}{signal-to-noise ratio}
\newacronym{rssm}{RSSM}{recurrent state-space model}
\newacronym{rl}{RL}{reinforcement learning}

\glsdisablehyper

\title{From Pixels to CSI: Distilling Latent Dynamics For Efficient Wireless Resource Management
\thanks{%
This work was supported in part by the ERA-NET CHIST-ERA Project MUSE-COM\textsuperscript{2}; in part by the Research Council of Finland Project Vision-Guided Wireless Communication; in part by the RCF-Korea Project Semantics-Native Communication and Protocol Learning in 6G; and in part by the European Union through the Project CENTRIC under Grant 101096379.}
}

\author{
\IEEEauthorblockN{%
Charbel Bou Chaaya, Abanoub M. Girgis and Mehdi Bennis
}
\IEEEauthorblockA{%
Centre for Wireless Communications, University of Oulu, Finland\\
Emails: \{charbel.bouchaaya, abanoub.pipaoy, mehdi.bennis\}@oulu.fi}
}

\maketitle


\begin{abstract}
In this work, we aim to optimize the radio resource management of a communication system between a remote controller and its device, whose state is represented through image frames, without compromising the performance of the control task.
We propose a novel \gls{ml} technique to jointly model and predict the dynamics of the control system as well as the wireless propagation environment in latent space.
Our method leverages two coupled \glspl{jepa}: a control \gls{jepa} models the control dynamics and guides the predictions of a wireless \gls{jepa}, which captures the dynamics of the device's \gls{csi} through cross-modal conditioning.
We then train a deep \gls{rl} algorithm to derive a control policy from latent control dynamics and a power predictor to estimate scheduling intervals with favorable channel conditions based on latent \gls{csi} representations.
As such, the controller minimizes the usage of radio resources by utilizing the coupled \gls{jepa} networks to imagine the device's trajectory in latent space.
We present simulation results on synthetic multimodal data and show that our proposed approach reduces transmit power by over 50\% while maintaining control performance comparable to baseline methods that do not account for wireless optimization.
\end{abstract}


\glsresetall

\section{Introduction}
\IEEEPARstart{T}{he} development of smart factories and the proliferation of smart devices entails a fundamental shift in optimizing wireless networks, in which use cases such as digital twins and the metaverse will strain the capacity of current networks.
This is mainly due to the intelligent nature of devices that no longer transmit passive data, but rather communicate computation models or offload their processing needs to solve tasks.
This comes in tandem with the accelerating progress in \gls{ai} and its remarkable impact on various layers of control and communication.
Hence, an intrinsic convergence of communication, control, and \gls{ai} is essential in the design of future wireless systems~\cite{park2022extreme}.

Recently, optimizing wireless networks to remotely control devices has gained significant attention in the literature.
Typically, sensors communicate their raw states to remote controllers equipped with computational resources, which in turn transmit actions, steering devices to achieve their objectives.
Current approaches aim to jointly design control and communication frameworks to solve tasks of devices while managing radio resources.
One such approach is to define freshness metrics reflecting the contextual importance of control systems as~\cite{cao2023age} and~\cite{kaul2011minimizing}.
Then, joint control and communication policies are derived by optimizing a trade-off between control stability and wireless resource utilization.
Another prominent approach is to equip a controller with a predictive model of the device's state, through which it infers the impact of its policy on the device.
For instance, Gaussian processes were employed in~\cite{girgis2021predictive} and generative adversarial networks in~\cite{kizilkaya2023task} to predict the device's state at the controller which significantly reduces communication overhead.
However, the control processes considered in these works are very simplified, assuming known control dynamics operating on raw device states.
Nevertheless, modern devices acquire their sensory data through high-dimensional observations such as images and point clouds~\cite{oneil2024rtx}.
Hence, optimizing such remote control systems is challenging under unknown dynamics and considerably large state observations, where computing freshness metrics or directly estimating the state become infeasible.

To overcome those limitations, we propose a novel data-driven framework where the controller learns the dynamics of the device's control environment captured by images, as well as the dynamics of the wireless environment through estimated \gls{csi}.
The goal is to minimize the usage of radio resources without compromising the device's control objective.
As both sensory observations are high-dimensional, the dynamics are learned in abstract latent space, using \glspl{jepa}~\cite{lecun2022path}.
Essentially, we couple a control \gls{jepa}, that simulates the latent control dynamics from pixels, with a wireless \gls{jepa} that learns the dynamics of the device's \gls{csi}.
We guide the prediction of the wireless dynamics by distilling its conditioning information from the latent control model.
As such, the control \gls{jepa} is utilized to predict the device's state, whereas the wireless \gls{jepa} determines well-chosen slots to receive the actual state with minimal use of radio resources.
The controller's action policy is derived by training a deep \gls{rl} algorithm on top of the control \gls{jepa}.
Our work draws inspiration from the model-based Dreamer agent, which was shown to solve complex and diverse synthetic and physical tasks~\cite{hafner2021mastering},~\cite{hafner2023mastering},~\cite{wu2023daydreamer}.

The only work comparable to ours is~\cite{girgis2024time}, which considers a vision-based remote control system, and uses control \gls{jepa} state predictions whenever communicated packets are dropped due to fading, maintaining appropriate control performance.
Our aim is to further relax physical resources under unknown control and wireless channel dynamics, which is not attempted in~\cite{girgis2024time}.
We present simulation results performed in a synthetic environment with multimodal data and show that our proposed method significantly minimizes radio resource usage, namely 50\% less transmit power compared to baselines, without impacting the control task.

\section{System Model}

\begin{figure}
    \centering
    \includegraphics[width=0.5\linewidth]{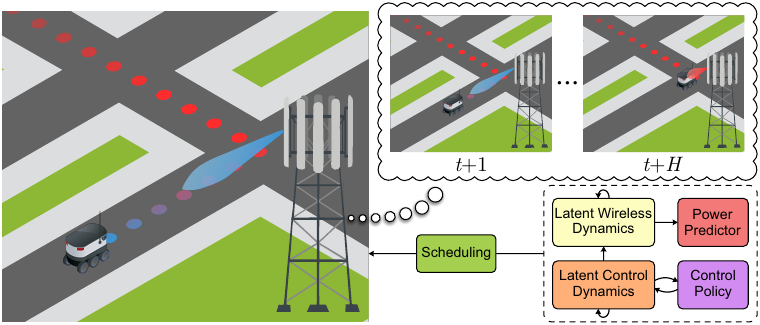}
    \caption{System model and solution scheme: the controller imagines the device's future multimodal states and optimizes its communication-control policy.}
    \label{fig:wi_dreamer_conf_system_model}
\end{figure}

We consider a time division duplex wireless network, where a base station (remote controller) equipped with $\numantennas$ antennas serves a single antenna sensor-actuator device, as shown in Fig.~\ref{fig:wi_dreamer_conf_system_model}.
The device's physical state is modeled as a closed-loop process, that must be controlled to solve a pre-defined task.

\subsection{Control Model}
In a control loop system, a sensor measures the device's state through images, typically acquired via a camera positioned above the device.
The actuator then applies a control action to steer the device to its desired state.
The device is modeled as a discounted \gls{mdp} with discrete time steps, defined by the tuple $\rbrk{\mdpstatespace, \mdpactionspace, \mdptransition, \reward, \mdpdiscount}$, where:
\begin{itemize}[nolistsep, leftmargin=*]
    \item $\mdpstatespace$ is the device's state space, i.e., set of images representing the device's state,
    \item $\mdpactionspace$ is a set of feasible actions, i.e., actuator set,
    \item $\mdptransition: \mdpstatespace \times \mdpactionspace \to \Delta\mdpstatespace$ is the unknown transition dynamics,
    \item $\reward: \mdpstatespace \times \mdpactionspace \to \mathbb{R}$ is a scalar reward function,
    \item $\mdpdiscount$ is a discount factor.
\end{itemize}
The actuator seeks actions from a policy \mbox{$\actionpolicy: \mdpstatespace \to \Delta\mdpactionspace$} that maximizes the expected sum of discounted rewards $\mathbb{E}\bigl[\sum_{t=0}^{\infty} \mdpdiscount^t \reward\rbrk{\state_t, \action_t}\bigr]$.
Note that unlike the predominant literature (e.g., \cite{girgis2021predictive}), we only assume access to images of the device as observations, rather than the device's raw physical state.
Under limited computational capacity, the device must communicate with the base station in order to solve the task.

\subsection{Communication Model}
The device's state is received by the remote controller in the uplink, while the downlink is dedicated to the controller to communicate the proposed actions.
Typically, since the control action's size is negligible compared to the state, we primarily focus our study on the wireless uplink.
At time slot $t$, we denote by $\channel_t \in \mathbb{C}^{\numantennas}$ the device's channel, which is estimated by the base station.
Since wireless resources must be shared by other services, the device's state cannot be always transmitted.
Hence, we further define a binary scheduling variable $\scheduling_t$ which indicates whether the base station decides to sample the device state or not.
When a communication is scheduled, the device transmits its state as a packet over the wireless channel with transmit power $\power_t$. 
Thus, the received signal-to-noise ratio can be written as: $\snr_t = \frac{\lvert\channel_t\rvert^2 \power_t}{\noisepower^2}$, where $\noisepower^2$ is the power of additive noise.
%


\subsection{Problem Formulation}
Given the above system, the device's high-dimensional pixel states cannot be continuously sent to the controller as this will incur considerable communication power and strain the network's capacity.
In this work, we aim to balance the trade-off between device control performance and the communication cost incurred in terms of power consumption.
We define the long-term averages of control reward and power utilization:
\begin{equation*}
    \longtermreward = \lim_{t\to\infty} \frac{1}{t} \sum_{t^\prime=1}^t \reward \rbrk{\state_{t^\prime}, \action_{t^\prime}},
    \quad
    \longtermpower = \lim_{t\to\infty} \frac{1}{t} \sum_{t^\prime=1}^t \scheduling_{t^\prime} \power_{t^\prime}.
\end{equation*}
The problem is formalized as follows:
\begin{subequations}\label{eq:wi_dreamer_conf_problem}
\begin{align}
    \label{eq:wi_dreamer_conf_objective} 
    \tag{\ref{eq:wi_dreamer_conf_problem}}
    &\underset{\rbrk{\scheduling_t, \, \action_t, \, \power_t}}{\text{maximize}} && \rbrk{\longtermreward, -\longtermpower} \\
    \label{eq:wi_dreamer_conf_constraint1}
    &\text{subject to}&& 0 \leq \scheduling_t \, \power_t \leq \powerbudget  && \forall\, t, \\
    \label{eq:wi_dreamer_conf_constraint2}
    &&& \snr_t \geq \scheduling_t \, \snrtarget && \forall\, t, \\
    \label{eq:wi_dreamer_conf_constraint3}
    &&& \scheduling_t \in \cbrk{0, 1}, \; \action_t \in \mdpactionspace && \forall\, t.
\end{align}
\end{subequations}
With the above multi-objective optimization in \eqref{eq:wi_dreamer_conf_objective}, we seek a joint communication-control policy that ensures the control task is executed by maximizing the long-term average reward $\longtermreward$, while minimizing the usage of wireless resources, i.e. the long-term average power $\longtermpower$.
When the device sends its state, the uplink transmit power is constrained by a given power budget $\powerbudget$ as shown in \eqref{eq:wi_dreamer_conf_constraint1}, while \eqref{eq:wi_dreamer_conf_constraint2} constrains the uplink $\snr$ to be above a threshold to ensure the state is decoded.
Finally, \eqref{eq:wi_dreamer_conf_constraint3} ensures the feasibility of the control action and the scheduling variable.

The aforementioned optimization problem is highly challenging due to the intricate coupling between control and communication variables.
In fact, while the controller requires fresh state observations to compute actions that drive the device to complete its task, such observations are costly in terms of wireless resources since they are captured by high-dimensional images, which necessitate abundant transmit power.
Moreover, the explicit coupling between the scheduling variable and the device's control state or action cannot be derived, as we make no assumption on the underlying device task, control dynamics, or wireless channel model, and only expect access to high-dimensional images as states.
Therefore, the base station must smartly schedule the device's transmissions by optimizing the usage of uplink resources, without compromising the control objective.

\section{Proposed Solution}
Solving problem~\eqref{eq:wi_dreamer_conf_problem} involves learning the device's transition dynamics $\mdptransition$.
Essentially, if the controller possesses a model of the control dynamics, it can considerably minimize the communication overhead with the device by anticipating the impact of the actions it sends in the downlink.
However, we recall that $\mdptransition$ is a generative model that outputs observations in pixel space, which are high-dimensional and contain redundant information (such as the color of certain objects in the environment or the position of some items in the background).
Hence, instead of modeling $\mdptransition$, we learn the system's dynamics in the space of image representations, i.e. encoding high-dimensional images into sufficient representations from which we predict appropriate action sequences and corresponding future image representations.
Namely, the controller must predict, during the time slots when the device is not scheduled, the system's dynamics in latent space.
Not only that, the control dynamics model must be coupled with a wireless dynamics model that infers, given the predicted device dynamics, favorable scheduling slots to receive the device state with minimal power.

Concretely, as we show in Fig.~\ref{fig:wi_dreamer_conf_system_model}, our solution involves four main components:
\begin{itemize}[nolistsep, leftmargin=*]
    \item Two coupled networks to predict the device's control and \gls{csi} dynamics:
    \begin{itemize}[nolistsep, leftmargin=*, label=\scalebox{.425}{$\blacksquare$}]
        \item Control \gls{jepa}: a network $\controlnetparameters$ modeling the device's control dynamics, by encoding its pixel states to abstract representations that are predictable given a sequence of control actions.
        \item Wireless \gls{jepa}: a network $\channelnetparameters$ modeling the device's channel dynamics, by embedding its \gls{csi} into a low-dimensional space, that are predictable given the device's control dynamics.
    \end{itemize}
    \item A control policy network $\actorparameters$, trained by observing the state's representations, that finds suitable control actions solving the device's task.
    \item A power prediction network $\powerpredictorparameters$ that observes the imagined \gls{csi} embeddings and infers the required transmit power for the device to send its state.
\end{itemize}
Equipped with those models, our strategy to solve~\eqref{eq:wi_dreamer_conf_problem} is the following:
The controller utilizes its \gls{jepa} networks to unroll the device's future multimodal states (for a given prediction horizon), hence evaluating its performance on the downstream task, as well as its estimated channel conditions.
With such predictions, the controller schedules the device's future transmission in the slot with the lowest estimated transmit power.
As such, since the control policy is trained in representation space, the controller utilizes its \gls{jepa} predictions to send appropriate actions in the downlink, and whenever the device is scheduled, the controller uses its fresh observations to update its model and facilitate future predictions.

To solve~\eqref{eq:wi_dreamer_conf_problem}, we propose to learn such models from data, and assume access to an experience dataset consisting of \gls{mdp} states as well as channel realizations.
In other words, the controller base station trains its models during an offline period prior to deployment, during which it always observes and learns to control the device.
All the mentioned networks are implemented by neural networks, whose training is detailed next.

\subsection{Learning Latent Control Dynamics from Pixels}
\begin{figure}[!t]
    \begin{minipage}[t]{.5\linewidth}
        \centering
        \includegraphics[width=\linewidth]{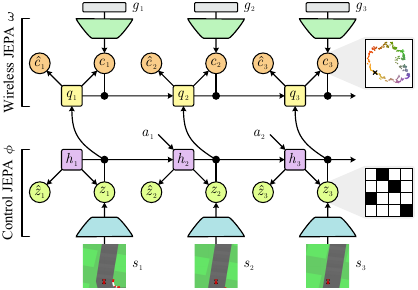}
        \caption{Learning latent control and wireless dynamics.}
        \label{fig:wi_dreamer_conf_architecture_train}
        \end{minipage}
    \hfill
    \begin{minipage}[t]{.5\linewidth}
        \centering
        \includegraphics[width=\linewidth]{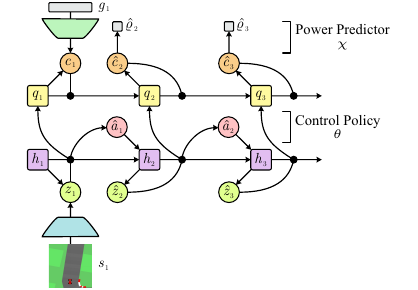}
        \caption{Learning control policy and power prediction in latent space.}
        \label{fig:wi_dreamer_conf_architecture_imagine}
    \end{minipage}
\end{figure}

Following Dreamer~\cite{hafner2023mastering}, we learn the control dynamics in latent space using a control \gls{jepa} composed of three networks: an image encoder, a \gls{rssm}~\cite{hafner2019learning}, and reward / termination predictors.
We present the model in Fig.~\ref{fig:wi_dreamer_conf_architecture_train}.
The latent control state is the concatenation of a deterministic variable $\deterministicstate_t$ and a stochastic variable $\stochasticstate_t$, which captures both aspects of the control transition.
All components are parametrized by a common set of learnable weights $\controlnetparameters$ and jointly trained.

The model training proceeds as follows.
The image encoder embeds visual observations to representations: $\imagerepresentation_t = \imageencoder_{\controlnetparameters}\rbrk{\state_t}$.
The \gls{rssm} is composed of three sub-networks.
First, a recurrent network (e.g. a GRU) that updates the deterministic hidden state \mbox{$\deterministicstate_{t} = \recurrentfunction_{\controlnetparameters}\rbrk{\deterministicstate_{t-1}, \action_{t-1}, \stochasticstate_{t-1}}$} given the previous action $\action_{t-1}$ and stochastic state $\stochasticstate_{t-1}$.
Second, a representation model that computes a posterior over the stochastic state \mbox{$\stochasticstate_t \sim \stochasticstatedistribution_{\controlnetparameters}\rbrk{\stochasticstate_t \,\vert\, \deterministicstate_t, \imagerepresentation_t}$} given the recurrent state and image features.
Third, a dynamics model that guesses the stochastic state \mbox{$\hat{\stochasticstate}_t \sim \distribution_{\controlnetparameters}\rbrk{\hat{\stochasticstate}_t \,\vert\, \deterministicstate_t}$} without accessing the image.
During training, the dynamics model is pushed to predict future model states given current model states and actions (as the recurrent state is deterministically updated), hence simulating the latent control dynamics captured by the representation model without observing the images.
Finally, given the latent state $\rbrk{\deterministicstate_t, \stochasticstate_t}$, the reward predictor infers the reward \mbox{$\hat{\reward}_t \sim \distribution_{\controlnetparameters}\rbrk{\hat{\reward}_t \,\vert\, \deterministicstate_t, \stochasticstate_t}$}, and the termination predictor \mbox{$\hat{\mdpdiscount}_t \sim \distribution_{\controlnetparameters}\rbrk{\hat{\mdpdiscount}_t \,\vert\, \deterministicstate_t, \stochasticstate_t}$} predicts whether the episode terminates (discount factor is set to zero for terminal steps).
We omit those two networks from Fig.~\ref{fig:wi_dreamer_conf_architecture_train} for clarity.

The stochastic variable $\stochasticstate_t$ is a collection of categorical variables, which has shown better results on control tasks compared to continuous variables.
The reward predictor outputs the mean of a unit variance Gaussian variable, and the termination predictor yields a Bernoulli variable.

Given a prediction horizon $\predictionhorizon$, the model parameters are optimized end-to-end to minimize the following loss:
\begin{align}
\begin{split}
    \label{eq:wi_dreamer_conf_control_loss} 
    \ell\rbrk{\controlnetparameters} = \mathbb{E}_{\stochasticstatedistribution_{\controlnetparameters}} \Bigr[ \textstyle{\sum}_{t=1}^\predictionhorizon \,
    \kllossscale \underbracket[0.100ex][0.300ex]{D_{\text{KL}} \sbrk{\stochasticstatedistribution_{\controlnetparameters}\rbrk{\stochasticstate_t \,\vert\, \deterministicstate_t, \imagerepresentation_t} \vert\vert \, \distribution_{\controlnetparameters}\rbrk{\stochasticstate_t \,\vert\, \deterministicstate_t}}}_{\text{KL loss}}
    \underbracket[0.100ex][0.300ex]{- \log\distribution_{\controlnetparameters}\rbrk{\reward_t \,\vert\, \deterministicstate_t, \stochasticstate_t}}_{\text{reward prediction loss}} 
    \underbracket[0.100ex][0.300ex]{- \log\distribution_{\controlnetparameters}\rbrk{\mdpdiscount_t \,\vert\, \deterministicstate_t, \stochasticstate_t}}_{\text{termination prediction loss}} \Bigl]
\end{split}
\end{align}
Essentially, the model is trained to produce representations from which the reward and termination networks predict the likelihood of their corresponding targets.
Further, the KL loss, scaled by $\kllossscale$, regularizes the representations as follows.
On the one hand, it minimizes the discrepancy between the dynamics predictor's output and the next representation.
On the other hand, it pushes the representation model to produce more predictable representations to facilitate the dynamics model's task.
One can think of the KL loss as the typical \gls{jepa} prediction loss, with the reward and termination prediction losses serving for further representation regularization.
Following Dreamer, to avoid degenerate representations from which the dynamics model trivially predicts future states, but do not contain sufficient information, we weight the KL term higher with respect to the representation model than the dynamics predictor as follows:
\begin{align}
\begin{split}
    \label{eq:wi_dreamer_conf_kl_loss} 
    \text{KL loss} = 
    \kllossfactor \, D_{\text{KL}} \sbrk{\stopgradient\rbrk{\stochasticstatedistribution_{\controlnetparameters}\rbrk{\stochasticstate_t \,\vert\, \deterministicstate_t, \imagerepresentation_t}} \vert\vert \, \distribution_{\controlnetparameters}\rbrk{\stochasticstate_t \,\vert\, \deterministicstate_t}}
    + \rbrk{1-\kllossfactor} \, D_{\text{KL}} \sbrk{\stochasticstatedistribution_{\controlnetparameters}\rbrk{\stochasticstate_t \,\vert\, \deterministicstate_t, \imagerepresentation_t} \vert\vert \, \stopgradient\rbrk{\distribution_{\controlnetparameters}\rbrk{\stochasticstate_t \,\vert\, \deterministicstate_t}}}
\end{split}
\end{align}
where $\stopgradient\rbrk{\cdot}$ is the stop-gradient operator, and $\kllossfactor$ is a hyperparameter.

Unlike Dreamer, we do not regularize the representation by reconstructing the original pixel observations, as this restricts modeling unnecessary information and might dismiss important aspects of the task's objective.
In order to prevent representation collapse, we apply a batch normalization layer~\cite{ioffe15} inside the representation model.
We found this to be sufficient for stable training in our case.
However, one might further regularize the reconstruction-free model using contrastive learning approaches as done in~\cite{paster2021blast} for instance.

Equipped with this \gls{jepa}, the controller can predict the device's control dynamics in latent space.
We now couple the learned control dynamics with the dynamics of the wireless environment.


\subsection{Learning Latent Wireless Dynamics from CSI with Cross-modal conditioning}
We are now interested in modeling the impact of the device's control on its \gls{csi}.
In practice, the user's channel dynamics is a function of its movement which is dictated by its control state.
Hence, one can think of inferring future \gls{csi} by conditioning on the device's predicted control state.
However, in our realistic case, the controller only accesses pixel frames as control states.
Accordingly, \emph{we propose to distill the latent control features learned from the pixel modality as a conditional information that guides the prediction over future wireless \gls{csi}}.
Likewise, instead of training a generative model that predicts future \gls{csi}, we focus on learning the channel's dynamics in latent space, which is easier and sufficient to characterize low transmit power slots.

This model, parametrized by learnable parameters $\channelnetparameters$, involves the following two networks, as shown in Fig.~\ref{fig:wi_dreamer_conf_architecture_train}.
A channel encoder $\channelencoder$ that maps high-dimensional channels to their latent representations $\channelrepresentation_t = \channelencoder_{\channelnetparameters}\rbrk{\channel_t}$, and a recurrent prediction network $\channelpredictor$, with hidden state $\channelpredictorhiddenstate$, that receives the encoder's output and guesses future embeddings given the latent control state $\hat{\channelrepresentation}_t = \channelpredictor_{\channelnetparameters} \rbrk{\channelrepresentation_{t-1} \,\vert\, \channelpredictorhiddenstate_t, \deterministicstate_t, \stochasticstate_t}$.
As such, the controller models the intricate coupling between the control state, observed as image frames, and the device's wireless channel in abstract latent space.

Our model is variation of the recently proposed wireless \gls{jepa}~\cite{chaaya2024learning}, wherein the prediction is conditioned on the device's raw velocity.
As the velocity is not available in our case, we offload such information from the control dynamics model, which learns equivalent transition dynamics in latent space.
Our \emph{latent cross-modal distillation} functions as an imputation process, where the necessary information to predict one modality is transferred from another.
We facilitate this transfer within the latent space of both modalities (pixel and \gls{csi}).

With a prediction horizon $\predictionhorizon$, the model is trained end-to-end with the following loss:
\begin{equation}
    \label{eq:wi_dreamer_conf_wireless_loss} 
    \ell\rbrk{\channelnetparameters} = \mathbb{E}_{\distribution_{\controlnetparameters}} \sbrk{\textstyle{\sum}_{t=1}^{\predictionhorizon} \left\lvert \hat{\channelrepresentation}_{t} - \channelrepresentation_{t} \right\rvert^2}.
\end{equation}
While the predictor observes the encoder's output, we compare its predictions with slightly distorted versions of the encoder's embeddings to avoid representation collapse.
The distorted representations are obtained from a network whose weights are updated by a \gls{ema} of the encoder's weights.
We note that while training this \gls{jepa} network, the control dynamics network's weights are frozen.

With the coupled \glspl{jepa}, the controller can imagine the devices' future multimodal states, control and wireless.
We now propose to learn a control policy on top of the control \gls{jepa}, and a power prediction network on top of the wireless \gls{jepa}.


\subsection{Learning Control Policy and Power Prediction from Latent Representations}
We now develop a deep \gls{rl} framework to learn the device's control policy by imagining trajectories in latent space, as shown in Fig.~\ref{fig:wi_dreamer_conf_architecture_imagine}.
By doing so, the controller can then relax communication overhead with the device, since it can predict the device's dynamics without sampling the state, while still sending convenient actions.
Unlike~\cite{girgis2024time} which assumes access to an optimal control policy over pixel observations which is used to fit an actor model from latent representations, we follow a more general task-agnostic \gls{rl} method.

We learn the control policy by cooperatively training an actor network and a critic network, parametrized by $\actorparameters$ and $\criticparameters$ respectively, as follows:
\begin{equation*}
    \label{eq:wi_dreamer_conf_actor_critic} 
    \text{Actor:} \; \hat{\action}_t \sim \actionpolicy_{\actorparameters} \rbrk{\hat{\action}_t \,\vert\, \hat{\stochasticstate}_t}, \quad
    \text{Critic:} \; \valuefunction_{\criticparameters}\rbrk{\hat{\stochasticstate}_t} \approx \mathbb{E}_{\distribution_{\controlnetparameters}, \actionpolicy_{\actorparameters}} \sbrk{\valuefunction_t^{\lambda}}.
\end{equation*}
The control dynamics model is used to imagine future latent trajectories given the stochastic actor's choices, while the critic collects the imagined discounted rewards.
Hence, while the critic estimates the returns achieved by the actor, the actor is trained to maximize the critic's output.
Note that both networks are trained from the imagined states and rewards of the control \gls{jepa}, which is fixed during this training.

Given an imagined trajectory of $\predictionhorizon$ steps, the critic is trained with the following loss:
\begin{equation}
    \label{eq:wi_dreamer_conf_critic_loss} 
    \ell\rbrk{\criticparameters} = \mathbb{E}_{\distribution_{\controlnetparameters}, \actionpolicy_{\actorparameters}} \sbrk{\textstyle{\sum}_{t=1}^{\predictionhorizon-1} \, \frac{1}{2}\rbrk{\valuefunction_{\criticparameters}\rbrk{\hat{\stochasticstate}_t} - \stopgradient\rbrk{\valuefunction_t^\lambda}}^2},
\end{equation}
where the target value function is:
\begin{equation}
    \label{eq:wi_dreamer_conf_value_function}
    \valuefunction_t^\lambda \!=\! \hat{\reward}_t \!+\! \hat{\mdpdiscount}_t \rbrk{\rbrk{1-\lambda} \valuefunction_{\criticparameters} \rbrk{\hat{\stochasticstate}_{t+1}} \!+\! \lambda \valuefunction^\lambda_{t+1}}, \;
    \valuefunction_\predictionhorizon^\lambda \!=\! \valuefunction_{\criticparameters} \rbrk{\hat{\stochasticstate}_{\predictionhorizon}},
\end{equation}
and $\lambda$ is a parameter weighing longer horizon returns exponentially less.

On the other hand, we train the actor to maximize the return prediction made by the critic.
For that, we use reinforce gradients~\cite{williams1992simple} that regulate the probability of drawing actions by their expected return values.
Since we consider discrete actions, the actor's loss is:
\begin{align}
\begin{split}
    \label{eq:wi_dreamer_conf_actor_loss}
    \ell\rbrk{\actorparameters} = \mathbb{E}_{\distribution_{\controlnetparameters}, \actionpolicy_{\actorparameters}} \Bigl[ \textstyle{\sum}_{t=1}^{\predictionhorizon}
    \underbracket[0.100ex][0.300ex]{-\log \actionpolicy_\actorparameters \rbrk{\hat{\action}_t \,\vert\, \hat{\stochasticstate}_t} \stopgradient\rbrk{\valuefunction_t^\lambda - \valuefunction_{\criticparameters}\rbrk{\hat{\stochasticstate}_t}}}_{\text{reinforce loss}} 
    \underbracket[0.100ex][0.300ex]{\vphantom{-\log \actionpolicy_\actorparameters \rbrk{\hat{\action}_t \,\vert\, \hat{\stochasticstate}_t} \stopgradient\rbrk{\valuefunction_t^\lambda - \valuefunction_{\criticparameters}\rbrk{\hat{\stochasticstate}_t}}}-\actorentropyscale \, \mathrm{H}\sbrk{\actionpolicy_\actorparameters \rbrk{\hat{\action}_t \,\vert\, \hat{\stochasticstate}_t}}}_{\text{entropy regularization}} \Bigr],
\end{split}
\end{align}
where we regularize the actor's entropy to encourage exploration.

Since the above \gls{rl} agent can act on the device's state from imagined dynamics predicted by the control \gls{jepa} whenever the device is unscheduled, the final ingredient of our model is a network that predicts favorable scheduling slots requiring low transmit power to receive the device's state.
To learn that, we use the latent wireless dynamics model, and train a power prediction network $\hat{\power}_\powerpredictorparameters\rbrk{\channelrepresentation_t}$ that outputs the power needed to transmit the device's state while meeting the required $\snr$.
The network's parameters $\powerpredictorparameters$ are trained to minimize the following loss:
\begin{equation}
    \label{eq:wi_dreamer_conf_power_loss}
    \ell\rbrk{\powerpredictorparameters} = \mathbb{E}_{\distribution_{\controlnetparameters}, \actionpolicy_{\actorparameters}} \sbrk{\textstyle{\sum}_{t=1}^{\predictionhorizon}\rbrk{\hat{\power}_\powerpredictorparameters\rbrk{\channelrepresentation_t} - \frac{\snrtarget \noisepower^2}{\lvert\channel_t\rvert^2}}^2}.
\end{equation}
In a nutshell, as shown in Fig.~\ref{fig:wi_dreamer_conf_architecture_imagine}, the proposed \gls{jepa} networks allow the controller to roll-out the device future multimodal states, following the \gls{rl} actions as a conditional variable for future control states, which in turn serve as a conditional variable for future \gls{csi} embedings, from which the controller predicts the necessary transmit power for each future slot.

\subsection{Overall Solution}
Finally, integrating all the proposed blocks, we solve \eqref{eq:wi_dreamer_conf_objective} as follows.
First, we feed the controller's (most recent) received device states to roll-out its future control and wireless states for a certain horizon $\predictionhorizon$, as shown in Fig.~\ref{fig:wi_dreamer_conf_architecture_imagine}.
Given the latent trajectories, the controller schedules the device next transmission at the slot with minimal transmit power estimated by the power predictor, disregarding the slots where constraints \eqref{eq:wi_dreamer_conf_constraint1} and \eqref{eq:wi_dreamer_conf_constraint2} are violated.
During the slots when the device is unscheduled, the \gls{rl} policy (operating on latent predicted states) is used to send proposed actions in the downlink.
When the device's transmission occurs, the controller receives the device's ground-truth states, embeds them with the proposed \glspl{jepa} to facilitate future predictions and the procedure is repeated.
As such, the controller uses its latent predictive models to steer the device towards completing its task, while relaxing the communication spectrum by scheduling the device in slots with good channel conditions, hence solving~\eqref{eq:wi_dreamer_conf_objective}.

Practically, sending one state observation might not be enough to capture higher order dynamics, thus, we introduce a parameter $\consecutivesamples$ representing the number of consecutive samples or slots to schedule the device.
Further, to prevent frequent scheduling when the device is in favorable channel conditions, we let $\truststeps$ be the number of initial steps where scheduling is omitted and predictive models are used.
We study the impact of those parameters in the following section.

\section{Simulation Results}
\subsection{Setting}
To validate our proposed algorithm, we created a pipeline between the \gls{rl} environment interface gym~\cite{towers2024gymnasium} and the ray-tracing software Sionna~\cite{hoydis2022sionna}.
We utilized the `Car Racing' gym environment, where the device is a robotic car that must be controlled to drive along a given track.
We then designed a Sionna scene where the device's position is replicated from gym to render its wireless channel.
We dedicate Appendix~\ref{appendix} to disclose our simulation hyperparameters.

\subsection{Discussion}
\begin{figure*}
\centering
\begin{minipage}{.3\textwidth}
      \centering
      \includegraphics[width=.8\textwidth]{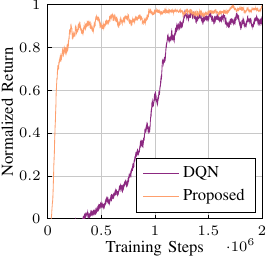}
      \caption{Convergence of \gls{rl} algorithms.}
      \label{fig:wi_dreamer_conf_rewards}
\end{minipage}%
\begin{minipage}{.7\textwidth}
    \centering\hspace{-6em}
    \subfloat[Ground truth locations.]{\parbox[t][3.8cm][t]{.42\textwidth}{
        \begin{tikzpicture}[baseline,remember picture]
    \begin{axis}[
        width=0.25\textwidth,
        height=0.25\textwidth,
        scale only axis,
        xmin=-60,
        xmax=60,
        ymin=-60,
        ymax=80,
        xlabel = {$x$ [m]},
        ylabel = {$y$ [m]},
        tick label style={font=\scriptsize},
        major tick length=2pt,
        every tick/.style={thin},
        label style={font=\footnotesize},
        ylabel shift = -8 pt,
        xlabel shift = -4 pt,
        xtick={-40, 0, 40, 80},
        ytick={-40, 0, 40},
        legend style={
          fill opacity=0.5,
          draw opacity=1,
          text opacity=1,
          at={(0.97,0.03)},
          anchor=south east,
          draw=black,
          font=\footnotesize
        },
        trajectory/.style={
          legend image code/.code={
            \foreach \x in {0,1,2,...,100}
                \draw[black!\x!red,only marks,mark=*,mark size=0.75,mark options={fill opacity=1}] plot coordinates {(1mm-\x/7,0cm)};
          }
        },
        antennas/.style={
          legend image code/.code={
            \draw[black,only marks,mark=triangle*,mark size=2,mark options={fill opacity=1,rotate=180}] plot coordinates {(1mm-50/7,0cm)};
          }
        }
    ]
        \addlegendimage{antennas}
        \addlegendimage{trajectory}
        \legend{Antennas, Trajectories}
        \addplot[thick,blue] graphics[xmin=-60,xmax=60,ymin=-60,ymax=80] {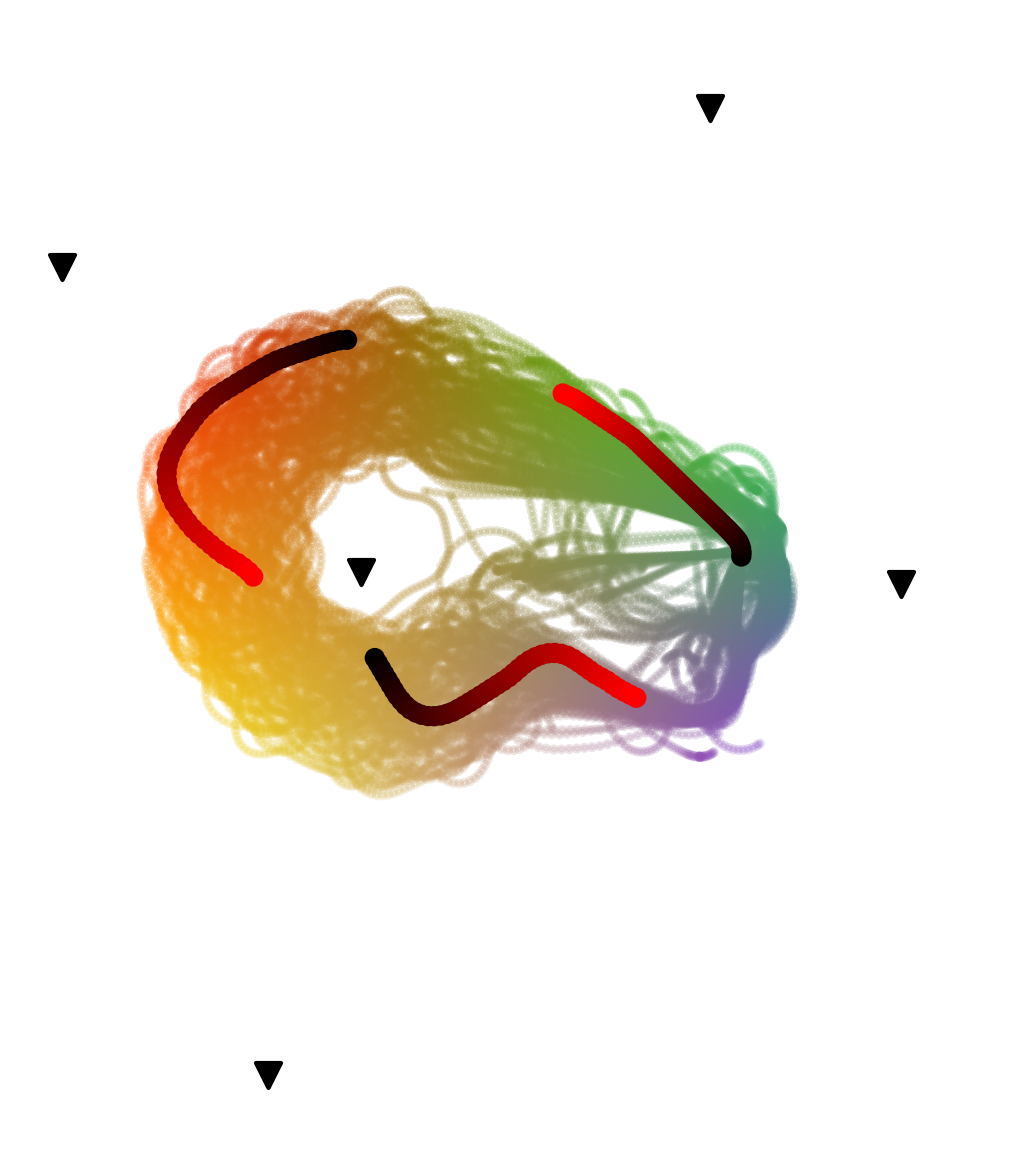};
    \end{axis}
\end{tikzpicture}
        \label{fig:wi_dreamer_conf_locations}}}
    \hspace{1em}
    \subfloat[Latent \gls{csi} space.]{\parbox[t][3.8cm][t]{.32\textwidth}{
        \begin{tikzpicture}[baseline,remember picture]
    \definecolor{pregreen}{RGB}{0,255,0}
    \definecolor{preblue}{RGB}{128,82,255}
    \begin{axis}[
        width=0.25\textwidth,
        height=0.25\textwidth,
        scale only axis,
        xmin=-200,
        xmax=200,
        ymin=-180,
        ymax=180,
        xlabel = {\phantom{$x$ [m]}},
        ylabel = {\phantom{$y$ [m]}},
        tick label style={font=\scriptsize},
        label style={font=\small},
        ylabel shift = -8 pt,
        xlabel shift = -4 pt,
        ticks=none,
        legend cell align=left,
        legend style={
          fill opacity=1,
          draw opacity=1,
          text opacity=1,
          at={(1.03,0.03)},
          anchor=south west,
          draw=black,
          font=\footnotesize
        },
        encoder/.style={
          legend image code/.code={
            \foreach \x in {0,1,2,...,100}
                \draw[black!\x!red,only marks,mark=*,,mark size=0.75,mark options={fill opacity=1}] plot coordinates {(1mm-\x/7,0cm)};
          }
        },
        predictor/.style={
          legend image code/.code={
            \foreach \x in {0,1,2,...,100}
                \draw[pregreen!\x!preblue,only marks,mark=*,,mark size=0.75,mark options={fill opacity=1}] plot coordinates {(1mm-\x/7,0cm)};
          }
        }
    ]
        \addlegendimage{encoder}
        \addlegendimage{predictor}
        \legend{Embeddings, Predictions}
        \addplot[thick,blue] graphics[xmin=-200,xmax=200,ymin=-180,ymax=180] {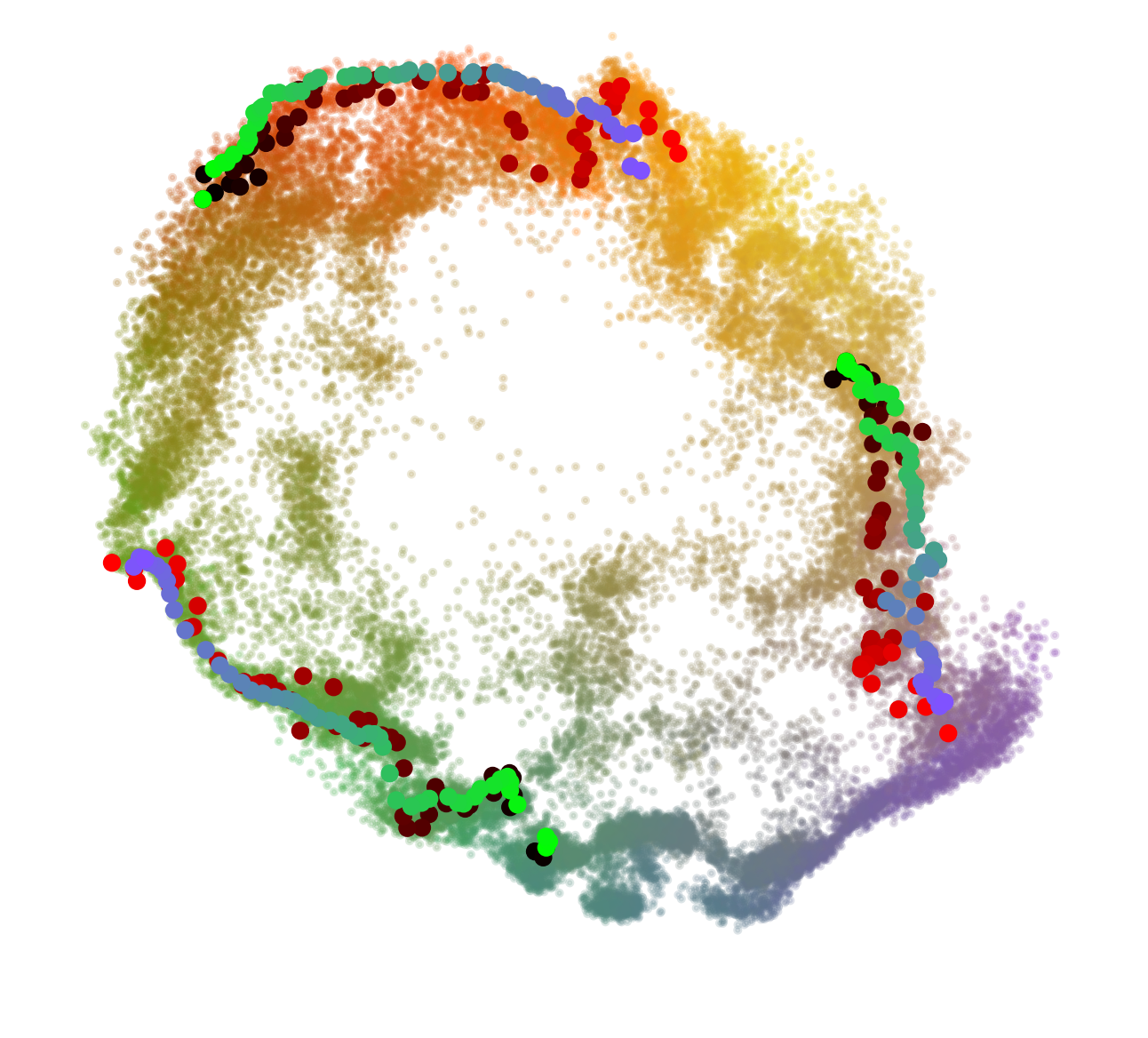};
    \end{axis}
\end{tikzpicture}
        \label{fig:wi_dreamer_conf_chart}}}
    \caption{Latent wireless dynamics learned by our model.}
    \label{fig:wi_dreamer_conf_cc}
\end{minipage}
\end{figure*}

In Fig.~\ref{fig:wi_dreamer_conf_rewards}, we show the convergence of our proposed \gls{rl} policy training algorithm compared to a model-free DQN~\cite{mnih2015human} baseline.
We notice that our algorithm converges faster reaching normalized rewards higher than $0.9$ after less than $0.5$ million epochs, which is achieved by the DQN agent after more than a million steps.
Both algorithms converge with similar returns around $0.93$ for the DQN method, and $0.98$ for our approach.
This confirms that our \gls{jepa} models learn sufficient state representations capturing only the important environment aspects (no representation collapse), which significantly simplifies the \gls{rl} training, and underscores the importance of learning the control policy from compact latent space instead of raw data.
We emphasize that the model-free benchmark cannot relax communication overhead as it does not learn the environment dynamics, rather only a policy over raw observations.

\begin{figure*}
    \centering
    \subfloat[Normalized return.]{
        \includegraphics{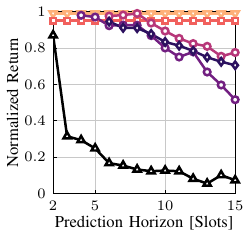}
        \label{fig:wi_dreamer_conf_comparison_score}}
    \hfill
    \subfloat[Transmit power.]{
        \includegraphics{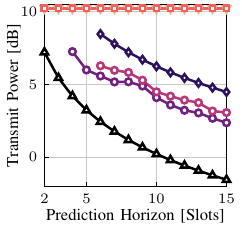}
        \label{fig:wi_dreamer_conf_comparison_power}}
    \hfill
    \subfloat[Communication overhead.]{
        \includegraphics{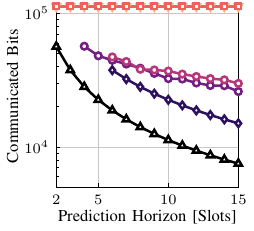}
        \label{fig:wi_dreamer_conf_comparison_bits}}
    \hfill
    \subfloat{\raisebox{2em}{
        \includegraphics{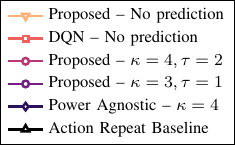}}}
    \caption{Comparison between different algorithms.}
    \label{fig:wi_dreamer_conf_comparison}
\end{figure*}

Fig.~\ref{fig:wi_dreamer_conf_locations} shows the ground-truth positions of the device and the base stations arrays, while Fig.~\ref{fig:wi_dreamer_conf_chart} presents the \gls{csi} embeddings as learned by our wireless \gls{jepa}.
We use gradient coloring to identify local neighborhoods.
We notice that the embeddings conserve the original environment's spatial structure, validating our idea of using latent control states to condition the wireless model's predictions.
Fig.~\ref{fig:wi_dreamer_conf_cc} further shows three particular trajectories taken by the car and their corresponding predictions made by our model from an initial channel estimation (black to red), against the actual \gls{csi} embeddings if the channels were estimated (green to blue).
We observe that our wireless \gls{jepa} accurately predicts future \gls{csi} embeddings in latent space, corroborating its robustness.

Fig.~\ref{fig:wi_dreamer_conf_comparison} plots the normalized control return, transmit power, and communication overhead for our method, compared to several baselines:
\begin{itemize}[nolistsep, leftmargin=*]
    \item No prediction: we use a variation of our model-based or a model-free approach where the controller always receives the state.
    \item Power agnostic: the controller utilizes its predictive control model and schedules the user at the end of the prediction horizon without considering power.
    \item An action repeat baseline: the controller with no predictive model receives one state per time horizon, and sends the same action for the rest of the time steps.
\end{itemize}
Focusing on our method, communicating $\consecutivesamples=4$ samples yields a higher return for a longer look-ahead horizon than the variant that communicates $\consecutivesamples=3$ samples.
For example, at $\predictionhorizon=10$ steps, the former achieves a normalized return of $0.9$, $15\%$ higher than the latter, while incurring only $0.4$ dB more transmit power on average.
We notice that our approach yields convenient control returns up to $10$-$12$ steps, allowing for significant spectrum relaxation, as the controller relies on its local models to steer the device with no communication.
Compared to the baselines with no prediction, our approach maintains a very similar control performance while saving $3$ times its transmit power and communicating less than $30\%$ in terms of bits with $10$ prediction slots.
We clearly observe that the action repeat benchmark communicates the least amount of data, but cannot achieve any significant return in terms of control.
Compared to the unimodal power agnostic control, our approach saves $3$ dB of transmit power ($50\%$ less) for short horizons and $2.2$ dB of transmit power ($40\%$ less) for longer horizons, while showcasing a similar control performance up to $10$ prediction steps with $\consecutivesamples=3$, and consistently outperforms it with the same number of consecutive samples $\consecutivesamples=4$.
We further notice that our methods communicate around $40\%$ more sample bits on average than the baseline with well-chosen slots (hence, less power), underscoring the importance of multi-modal designs.

\section{Conclusion}
In this work, we propose a novel self-supervised learning approach to predict the joint control and wireless dynamics in latent space and optimize wireless resources without compromising the control objective.
We employed two coupled \glspl{jepa}, one that simulates the control dynamics, and guides the prediction of wireless dynamics by the other \gls{jepa} through cross-modal conditioning.
We then trained a deep \gls{rl} algorithm to learn a control policy from the latent dynamics, and a power predictor that estimates scheduling intervals with good channel conditions from the latent \gls{csi} space.
Extensive results in a synthetic environment, reveal the efficiency and trade-offs of our proposed model.
Extensions of our work will include networks with multiple control systems sharing limited radio resources, possibly with correlated tasks.

\bibliography{references}
\bibliographystyle{IEEEtran}

\newpage

\appendices
\section{Hyperparameters}\label{appendix}

\subsection{Simulation Pipeline}

\begin{figure}[!h]
    \centering
    \includegraphics[width=0.5\linewidth]{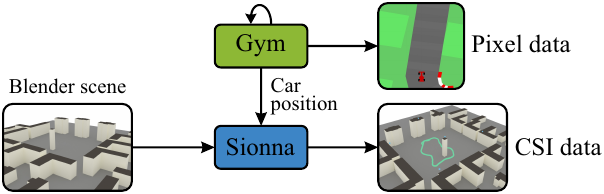}
    \label{fig:enter-label}
\end{figure}

\subsection{General Hyperparameters}

\begin{table}[H]
\centering
\caption{General Hyperparameters}
\begin{tabularx}{.5\linewidth}{
  >{\raggedright\arraybackslash}m{.2\linewidth} 
  >{\centering\arraybackslash}m{.3\linewidth}}
    \toprule
    Parameter & Value \\
    \midrule
    Optimizer & Adam \\
    Prediction horizon $\rbrk{H}$& $50$ \\
    Input image $\rbrk{s_t}$ & $84\times84$ grayscale \\
    Carrier frequency & $2.14$ GHz \\
    Subcarriers & $16$ \\
    Bandwidth & $20$ MHz \\
    Base station & $5$ arrays of $4 \times 2$ antennas \\
    Environment steps per gradient step & $20$ \\
    \bottomrule
\end{tabularx}
\end{table}

\subsection{Control JEPA Hyperparameters}

\begin{table}[H]
\centering
\caption{Control JEPA Hyperparameters}
\begin{tabularx}{.5\linewidth}{
  >{\raggedright\arraybackslash}m{.2\linewidth} 
  >{\centering\arraybackslash}m{.3\linewidth}}
    \toprule
    Parameter & Value \\
    \midrule
    Learning rate & $2 \times 10^{-4}$ \\
    Batch size & $32$ \\
    Gradient clipping & $100$ \\
    Latent variable $\rbrk{z_t}$ & $32$ categoricals with $32$ classes \\
    Discount factor $\rbrk{\gamma}$ & $0.99$ \\
    KL loss scale $\rbrk{\beta}$ & $0.5$ \\
    KL balacing $\rbrk{\mu}$ & $0.8$ \\
    Replay memory size & $10^{6}$ \\
    \bottomrule
\end{tabularx}
\end{table}

\subsection{Wireless JEPA Hyperparameters}

\begin{table}[H]
\centering
\caption{Wireless JEPA Hyperparameters}
\begin{tabularx}{.5\linewidth}{
  >{\raggedright\arraybackslash}m{.2\linewidth} 
  >{\centering\arraybackslash}m{.3\linewidth}}
    \toprule
    Parameter & Value \\
    \midrule
    Learning rate & $5 \times 10^{-3}$ \\
    Learning rate decay & $0.97$ \\
    Batch size & $100$ \\
    EMA decay & $0.99$ \\
    Weight decay & $3 \times 10^{-3}$ \\
    \bottomrule
\end{tabularx}
\end{table}

\subsection{RL Hyperparameters}

\begin{table}[H]
\centering
\caption{RL Hyperparameters}
\begin{tabularx}{.5\linewidth}{
  >{\raggedright\arraybackslash}m{.2\linewidth} 
  >{\centering\arraybackslash}m{.3\linewidth}}
    \toprule
    Parameter & Value \\
    \midrule
    Actor learning rate & $4 \times 10^{-5}$ \\
    Critic learning rate & $10^{-4}$ \\
    Return weight $\rbrk{\lambda}$ & $0.95$ \\
    Actor entropy scale $\rbrk{\eta}$ & $10^{-3}$ \\
    Slow target update & $1500$ steps \\
    \bottomrule
\end{tabularx}
\end{table}

\subsection{Neural Networks}
\subsubsection{Image encoder}
Three convolutional layers with $(16, 32, 64)$ channels, kernels $(8, 4, 2)$ and stride $(4, 2, 2)$, followed by two linear layers with $(1024, 256)$ neurons, with output size of $400$. All layers are followed by a layer normalization and ELU activation.
\subsubsection{RSSM recurrent network}
A GRU with a hidden state $h_t$ of size $300$. Its input $(a_{t-1}, z_{t-1})$ are fed to a linear layer with $300$ neurons followed by a layer normalization and ELU activation.
\subsubsection{RSSM representation network}
A linear layer with $400$ neurons that receives the image features and the recurrent state $(x_t, h_t)$ followed by a batch normalization and ELU activation, then another linear layer with $400$ neurons that outputs the $32 \times 32$ logits of the stochastic state $z_t$.
\subsubsection{RSSM dynamics network}
A linear layer with $400$ neurons that receives the recurrent state $h_t$ followed by a layer normalization and ELU activation, then another linear layer with $400$ neurons that outputs the $32 \times 32$ logits of the stochastic state $z_t$.
\subsubsection{Reward/Termination prediction networks}
Three layer MLPs with $100$ neurons per layer, and each layer is followed by a layer normalization and ELU activation.
\subsubsection{Actor/Critic network}
Three layer MLP with $100$ neurons per layer, and each layer is followed by a layer normalization and ELU activation.
\subsubsection{Channel encoder network}
Five layer MLP with $(1024, 512, 256, 128, 64)$ neurons, and each layer is followed by a batch normalization and ReLU activation.
\subsubsection{Channel prediction network}
A GRU with a recurrent state size of $256$ and its output is fed to two linear layers with $(64, 16)$ neurons.
\subsubsection{Power prediction network}
Three layer MLP with $100$ neurons per layer, and each layer is followed by a ReLU activation.



\end{document}